\documentclass[letterpaper, 10 pt, conference]{ieeeconf}  

\IEEEoverridecommandlockouts                              

\usepackage{booktabs}

\usepackage[noadjust]{cite}
\nocite{*}  

\makeatletter
\let\NAT@parse\undefined
\makeatother

\usepackage{hyperref}
\usepackage{balance}
\usepackage{comment}
\hypersetup{
    colorlinks=true,
    citecolor=Green,
    linkcolor=NavyBlue,
    filecolor=OrangeRed,      
    urlcolor=OrangeRed,
    pdfpagemode=FullScreen,
}

\usepackage{balance}
\usepackage{graphicx}
\usepackage{hyperref}
\usepackage{keyval}
\usepackage{algorithm}
\usepackage{algpseudocode}
\usepackage{amsmath, amsfonts}
\usepackage{amssymb}
\usepackage{cleveref}
\usepackage{fancyhdr} 
\usepackage{color}
\usepackage{transparent}
\usepackage{booktabs}
\usepackage{tabularx}

\usepackage{caption}
\usepackage{subcaption}
\usepackage{textcomp}
\usepackage{stfloats}
\usepackage{url}
\usepackage{verbatim} 
\usepackage{graphicx}
\usepackage[table, dvipsnames]{xcolor} 
\usepackage{multirow}
\usepackage{mathtools}
\let\labelindent\relax
\usepackage{enumitem} 
\usepackage{booktabs}

\usepackage{caption}
\usepackage{tabularx}
\usepackage{threeparttable}
\usepackage{makecell}

\newif\ifshowrevs
\showrevstrue

\newif\ifshowauthorcomment
\showauthorcommenttrue

\newif\ifkeep
\keepfalse

\newif\ifremvspace
\remvspacetrue

\newcommand{\idest}{i.e., }
\newcommand{\exempli}{e.g., }

\newif\ifshowstatshorizontal
\showstatshorizontaltrue

\newif\ifshowauthorcomment
\showauthorcommenttrue

\newcommand{\cut}[1]{}

\ifshowrevs

\else

\fi

\usepackage{xargs} 
\usepackage[colorinlistoftodos,prependcaption,textsize=tiny]{todonotes}
\newcommandx{\unsure}[2][1=]{\todo[linecolor=red,backgroundcolor=red!25,bordercolor=red,#1]{#2}}
\newcommandx{\change}[2][1=]{\todo[linecolor=blue,backgroundcolor=blue!25,bordercolor=blue,#1]{#2}}
\newcommandx{\info}[2][1=]{\todo[linecolor=OliveGreen,backgroundcolor=OliveGreen!25,bordercolor=OliveGreen,#1]{#2}}
\newcommandx{\improvement}[2][1=]{\todo[linecolor=Plum,backgroundcolor=Plum!25,bordercolor=Plum,#1]{#2}}
\newcommandx{\thiswillnotshow}[2][1=]{\todo[disable,#1]{#2}}
 
\def\HiLiYellow{\leavevmode\rlap{\hbox to \hsize{\color{yellow!20}\leaders\hrule height .8\baselineskip depth .4ex\hfill}}}
  
\fancypagestyle{firststyle}
{
  \fancyhf{}

}

\fancypagestyle{secondstyle}
{
  \fancyhf{}
  \fancyhead[R]{\footnotesize \thepage}

}

\usepackage{cleveref}

\graphicspath{{figures/}}

\title{\LARGE \bf
LongComp: Long-Tail Compositional Zero-Shot  \\ Generalization for Robust Trajectory Prediction
}

\author{Anonymous Submission for Double Blind Review
}

\author{Benjamin Stoler$^{1, 2}$ \and Jonathan Francis$^{1, 3}$ \and Jean Oh$^{1}$
\thanks{$^{1}$School of Computer Science, Carnegie Mellon University, Email: {\tt\footnotesize \{bstoler, jmf1, jeanoh\}@cs.cmu.edu}.
$^{2}$Stack AV Research Internship.
$^{3}$Bosch Center for Artificial Intelligence.}
}

\begin{document}

\maketitle
\thispagestyle{empty}
\pagestyle{empty}

\begin{abstract}

Methods for trajectory prediction in Autonomous Driving must contend with rare, safety-critical scenarios that make reliance on real-world data collection alone infeasible. To assess robustness under such conditions, we propose new long-tail evaluation settings that repartition datasets to create challenging out-of-distribution (OOD) test sets. We first introduce a safety-informed scenario factorization framework, which disentangles scenarios into discrete ego and social contexts. Building on analogies to compositional zero-shot image-labeling in Computer Vision, we then hold out novel context combinations to construct challenging closed-world and open-world settings. This process induces OOD performance gaps in future motion prediction of 5.0\% and 14.7\% in closed-world and open-world settings, respectively, relative to in-distribution performance for a state-of-the-art baseline. To improve generalization, we extend task-modular gating networks to operate within trajectory prediction models, and develop an auxiliary, difficulty-prediction head to refine internal representations. Our strategies jointly reduce the OOD performance gaps to 2.8\% and 11.5\% in the two settings, respectively, while still improving in-distribution performance.

\end{abstract}



\begin{figure*}[t]
  \centering
  \vspace{0.3cm}
  \begin{subfigure}{\linewidth}
    \centering
    \hspace*{3em}
    \includegraphics[width=0.7\linewidth]{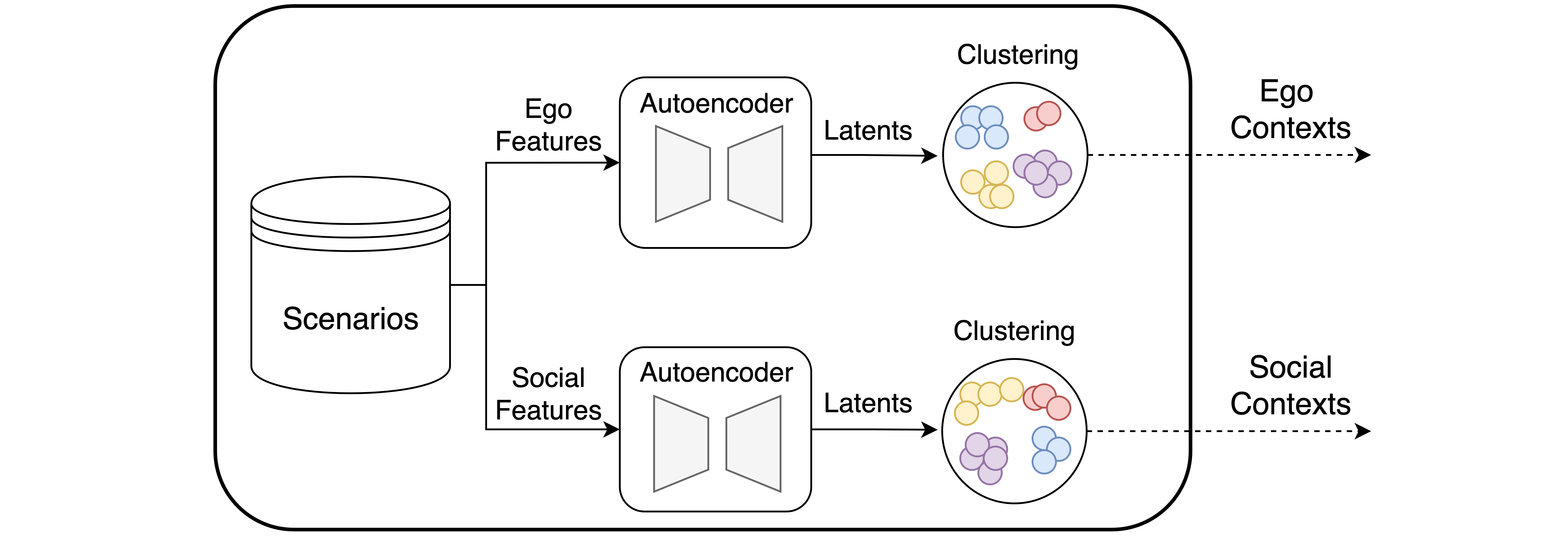}
    \caption{Scenario factorization into paired ego and social contexts.}
    \label{fig:overview_top}
  \end{subfigure}
  
  \vspace{0.3cm}
  
  \begin{subfigure}{\linewidth}
    \centering
    \includegraphics[width=0.9\linewidth]{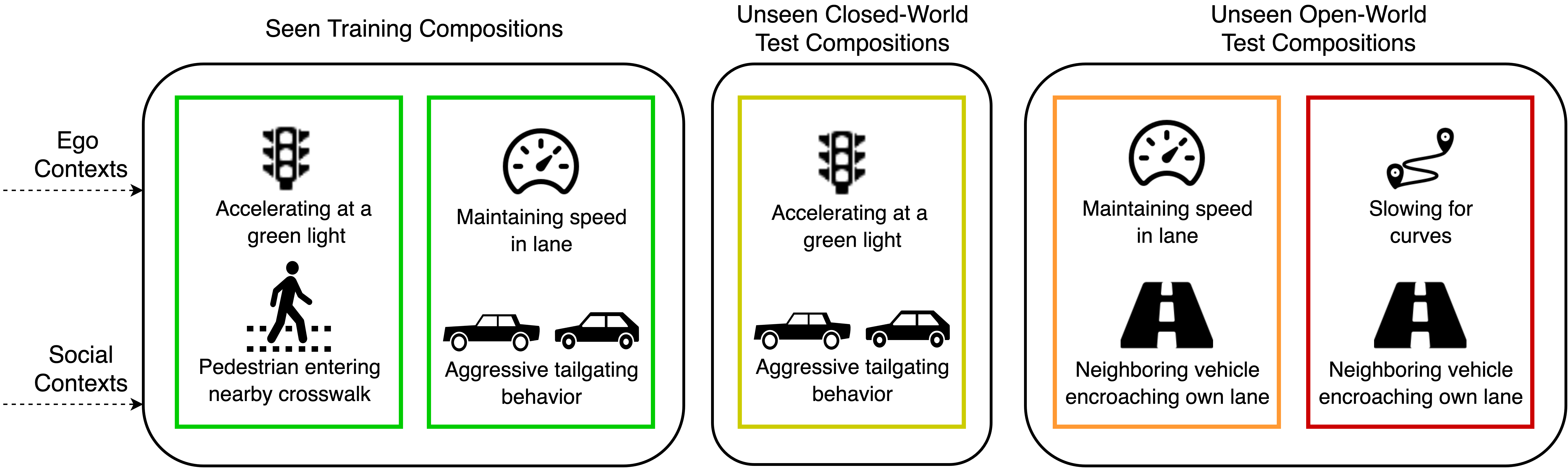}
    \caption{Illustrative toy examples of context compositions.}
    \label{fig:overview_bottom}
  \end{subfigure}

  \caption{Overview of our framework. (a) Traffic scenarios are factorized and clustered along explicitly disentangled ego and social axes. (b) These contexts are then used to create challenging compositional zero-shot evaluation settings for trajectory prediction, and to enable generalization strategies to enhance OOD robustness.}
  \label{fig:overview}
  
  \vspace{-0.5cm}
\end{figure*}


\section{Introduction} \label{sec:introduction}
In autonomous driving (AD), the long-tail distribution of safety-critical scenarios makes it infeasible to develop and assess systems by relying on large-scale, real-world data collection~\cite{liu2024curse}. Understanding and improving model performance in out-of-distribution (OOD) deployment settings has thus become increasingly important, including in the essential motion prediction component~\cite{wang2023bridging, ding2023survey}. Recent work has turned to creating artificial distribution shifts to emulate such OOD deployment settings, repartitioning existing datasets into non-uniform train and test sets on the bases of \exempli safety-relevance scores~\cite{stoler2024safeshift} or cluster identities~\cite{ye2024imgtp}. However, these bases are often highly entangled (\idest single axes that conflate many underlying attributes), which results in interpretation and generalization challenges, as emphasized in the broader machine learning literature~\cite{bengio2013representation, wang2024disentangled, lake2017building}.

Recently, work in related machine learning tasks, such as labeling objects in images, has developed a relevant framework known as compositional zero-shot learning (CZSL)~\cite{pourpanah2022review, mancini2021open}. In CZSL, object labels in images consist of pairs of object types and attributes. The zero-shot challenge is to train labeling models on some in-distribution (ID) \textit{seen} subset of these pairs and test on both seen and OOD \textit{unseen} pairs. These settings and resulting methodologies have demonstrated finer-grained evaluation of capabilities and improved model generalizability~\cite{atzmon2020causal, mancini2021open}. 
It is thus appealing to extend the CZSL setting to AD tasks, especially given both the hierarchical manner in which humans operate vehicles~\cite{medeiros2014hierarchical}, as well as the causal factorization of crash risks therein~\cite{stanton2009human, charlton2014risk}. In the context of driving, however, behavior generalization along safety-relevant, semantically meaningful axes has not yet been explored.

To address this gap, we propose and assess a safety-informed scenario factorization approach, enabling both the creation of challenging OOD evaluation settings and the development of robust, generalization methods for AD motion prediction. Existing studies on human driver error and risk assessment have organized contributing factors into pertinent categories: road infrastructure (road layout, signage quality), vehicle-related issues (mechanical reliability, maintenance), road user conditions (driver experience, mental state), behaviors of other road users (aggression, erraticism), and environmental conditions (lighting, weather)~\cite{stanton2009human, charlton2014risk}. Unfortunately, large motion prediction datasets typically lack much of this detailed information~\cite{ettinger2021large, wilson2023argoverse}.

We thus derive two axes from available indicators: an ``ego'' context, capturing both \textit{general} and \textit{safety-relevant} factors (such as kinematics, map features, and deviation from the speed limit), and a ``social'' context, capturing relative kinematics as well as safety-criticality indicators (including closing speeds and minimum time-to-conflict point difference (mTTCP)), as shown in \Cref{fig:overview_top}. 
We then create long-tail, zero-shot compositional closed-world and open-world settings on top of these paired contexts, with these axes serving as analogues to the object and attribute axes in the image-labeling CZSL setting.
These settings entail splitting data into seen and unseen portions non-uniformly, holding out novel combinations of ego and social contexts to be tested on, as shown in \Cref{fig:overview_bottom}. Note that, unlike the established image-labeling setting, our evaluation focuses on downstream task performance \textit{within} these novel context combinations, rather than predicting a label for a scenario; to our knowledge, this is the \textbf{first} extension of CZSL concepts to trajectory prediction.

In these closed-world and open-world settings, we observe a significant OOD performance gap in trajectory prediction, when using WOMD~\cite{ettinger2021large} as an exemplar dataset and MTR~\cite{shi2022motion} as a baseline prediction approach.
To enhance OOD performance, we then develop and extend domain generalization techniques from the image-labeling CZSL setting. In particular, we adapt task modular gating networks~\cite{purushwalkam2019task} to operate directly in the bottleneck layer of baseline approaches and further enhance the intermediate representation with an auxiliary, difficulty-prediction head.

Our contributions are thus as follows: 
\begin{enumerate}
    \item We introduce a novel, safety-informed scenario factorization approach for autonomous driving, leveraging explicitly disentangled ``ego'' and ``social'' axes.
    \item We propose new long-tail, zero-shot closed-world and open-world generalization settings upon these axes, where the difficulty-balanced prediction error in the closed-world test setting increases by an average of $5.0\%$ (closed-world) and $14.7\%$ (open-world) over their respective in-distribution performance.
    \item We develop generalization techniques that together reduce these OOD performance gaps to $2.8\%$ and $11.5\%$ respectively, eliminating nearly half the gap in closed-world and one quarter in open-world settings, while improving ID performance by $4.0\%$ and $1.2\%$.
\end{enumerate}

\section{Related Work} \label{sec:related_works}

\subsection{Compositional Zero-Shot Learning}

Compositional zero-shot learning (CZSL) is increasingly used in computer vision image-labeling as a framework for evaluating human-like generalization to out-of-distribution examples~\cite{pourpanah2022review}. In the canonical CZSL setting, models are tasked with labeling novel combinations of semantic factors (\idest object-attribute pairs) not observed during training~\cite{mancini2021open}. To reduce the generalization gap incurred, various strategies, such as task modular architectures~\cite{purushwalkam2019task}, conditional attribute learning~\cite{wang2023learning}, and attention propagation~\cite{khan2023learning}, have demonstrated promising performance. More recently, approaches leveraging large-scale pre-trained foundation models such as CLIP~\cite{nayak2023learning, bao2024prompting}, as well as retrieval-augmented generation (RAG)~\cite{jing2024retrieval}, have achieved substantial improvements therein.

In parallel, these evaluation settings and generalization strategies have also recently been extended from static image-labeling to video \textit{action}-labeling, with verb prompts serving as an analogue to attributes~\cite{li2024c2c, ye2025zero}. Nevertheless, our focus on both safety-relevant, semantically meaningful factorizations, as well as behavior \textit{generation} under these conditions rather than just labeling, remains largely unexplored. Furthermore, foundation models are less directly applicable in AD, where the heavy-tailed nature of driving behaviors makes large-scale pre-training alone less effective, especially in distribution shift settings~\cite{liu2024curse, dai2025large}.

\subsection{Scenario Characterization and AD Evaluation}

For rigorously developing AD systems, scenario characterization and mining approaches are essential, enabling efficient and effective training protocols and more structured evaluation procedures.
Identifying scenarios featuring high levels of interaction between traffic participants is particularly common in large dataset curation~\cite{ettinger2021large, wilson2023argoverse, moers2022exid}, ensuring that training and evaluation examples are sufficiently challenging. Recent approaches have expanded on this to introduce additional nuanced scenario characteristics, such as inter-agent causality~\cite{roelofs2022causalagents}, surprise potential~\cite{ding2025surprise}, and calibrated regret~\cite{nakamura2024general}. 
UniTraj~\cite{feng2024unitraj} further proposes a stratified evaluation procedure along Kalman-difficulty bins and agent trajectory shapes. Other approaches have explicitly focused on safety-relevance through automated analyses to capture near-critical scenarios, even while truly safety-critical, long-tail scenarios are absent from recorded datasets~\cite{glasmacher2022automated, stoler2024safeshift}. While much of this mining has historically been performed heuristically or with clustering, recent approaches have also explored using large language models and vision language models to identify challenging scenarios and obtain structured scenario descriptors~\cite{yang2024hard, tan2023language, zheng2024large}.

Beyond scenario understanding, non-uniform training and testing procedures have also been increasingly utilized. Prior work has explored creating artificial distribution shifts on various bases, including road geometry~\cite{filos2020can}, clustered ``domains'' stemming from agent trajectory and map statistics~\cite{ye2023improving, ye2024imgtp}, and overall safety-relevance~\cite{stoler2024safeshift}. Although some prior work, like \cite{wang2024drive}, has studied compositional OOD generalization along axes like weather and time-of-day, these surface-level axes do not consider the diverse behavior-level attributes that make AD uniquely challenging. Concurrently, many works have explored reducing the observed drops in performance, through techniques like Frenet-based domain normalization~\cite{ye2023improving}, collision avoidance losses~\cite{stoler2024safeshift}, and closed-loop training with synthetic generated scenarios~\cite{zhang2023cat, stoler2024seal}. However, the entangled nature of such prior domain split techniques makes both performance analysis and broader generalization challenging, with remediation techniques often tailored to specific settings. We instead disentangle scenarios along meaningful, safety-informed axes (\idest ego and social contexts), and extend modular CZSL techniques to target the gap arising from long-tail compositional challenges.

\section{Preliminaries} \label{sec:preliminaries}

In this section, we define relevant notation and task definitions for compositional zero-shot evaluation and generalization in trajectory prediction. Concretely, we utilize our pipeline in \Cref{ssec:characterization} and \Cref{ssec:discretization} to produce discrete ``ego'' and ``social'' context labels for agents, though the definitions and notations here are independent of those details.

\noindent \textbf{Scenario and Data Format:}
Given a particular dataset comprising the set of scenarios $\mathcal{S}$, we denote $\mathbf{s} \in \mathcal{S}$ to refer to a particular scenario therein. Each scenario $\mathbf{s}$ contains agent information for all traffic participants involved, denoted $\mathbf{X}$, as well as map and meta information, denoted $\mathbf{M}$.

We adopt the unified data format and features defined in UniTraj~\cite{feng2024unitraj}. That is, for a particular agent with id $i$, $\mathbf{X}_{i} \in \mathbf{X}$ is a time-indexed feature vector over $T$ timesteps, where $\mathbf{X}_i^{(t)}$ denotes agent $i$'s data at a particular timestep $t$. This information contains ground-plane kinematics (position, velocity, and acceleration), a valid bit indicating whether the agent is present, as well as one-hot encodings both of the agent's type and the current timestep; \idest an enriched trajectory representation. Information in $\mathbf{M}$ consists of polyline representations of road and traffic control devices, with locations interpolated at a fixed distance interval and a one-hot encoding of the feature type (\exempli lane, stop-sign, crosswalk, etc.), along with other relevant meta information (\exempli speed limit of lanes, lane adjacency graphs, etc.). Finally, $\mathbf{M}$ also contains a list of ``focal'' agents that must be motion-predicted, a subset of the total agents in $\mathbf{s}$.

\noindent \textbf{Compositional Zero-Shot Trajectory Prediction:}
In the standard motion prediction task, time-varying features in $\mathbf{s}$ are split into a \textit{history} and \textit{future}. Given $T=T_\text{hist} + T_\text{fut}$ representing the total number of timesteps in the scenario, we define $\mathbf{X}_{i}^{\text{hist}} = \{\mathbf{X}_i^{(t)}\}_{t=1}^{T_\text{hist}}$ as the history portion and $\mathbf{X}_{i}^{\text{fut}} = \{\mathbf{X}_i^{(t)}\}_{t=T_\text{hist}+1}^{T}$ as the future portion. The motion prediction task is then to predict the future ground-plane positions in $\mathbf{X}_{i}^{\text{fut}}$ given only $\mathbf{X}_{i}^{\text{hist}}$ and $\mathbf{M}$.

For a particular evaluation scheme, the total set of all agents must be split to form a training, validation, and test set. In the prototypical setting, these agents are split uniformly at random according to some desired size of each set, often at the scenario level of granularity (\idest all focal agents  within a given $\mathbf{s} \in \mathcal{S}$ are assigned to the same set). In our CZSL setting, we instead create splits at the agent level, as different focal agents within the same $\mathbf{s}$ may experience very different contexts. We first obtain paired categorical ego and social context labels from all agents across $\mathcal{S}$. We denote these contexts as $\mathcal{C_\text{ego}} = \{c_e^k\}_{k=1}^{N_\text{ego}}$ and $\mathcal{C_\text{social}} = \{c_s^k\}_{k=1}^{N_\text{social}}$ for some finite sizes $N_\text{ego}$ and $N_\text{social}$. Thus the compositional class set is defined as the Cartesian product $\mathcal{C} = \mathcal{C_\text{ego}} \times \mathcal{C_\text{social}}$, following the analogous formulation for image-labeling in \cite{jing2024retrieval}. This set is then divided into two subsets, $\mathcal{C^\text{seen}}$ for the known compositions, which make up the training and validation set, and $\mathcal{C^\text{unseen}}$ for the unknown compositions that make up the test set. Importantly, there is no overlap between $\mathcal{C^\text{seen}}$ and $\mathcal{C^\text{unseen}}$, meaning that agent examples in the test set are out-of-distribution with respect to the training set.

In the \textit{closed-world} test setting, $\mathcal{C^\text{unseen}}$ consists solely of novel \textit{combinations} of known ego and social contexts; that is, for each paired $(c_e,c_s)\in \mathcal{C^\text{unseen}}$, both $c_e$ and $c_s$ are present in $\mathcal{C^\text{seen}}$, but never jointly on the same example. In the harder \textit{open-world} test setting, however, we require that for each joint $(c_e, c_s) \in \mathcal{C^\text{unseen}}$, at least one of $c_e$ or $c_s$ is completely absent from $\mathcal{C^\text{seen}}$. 
Thus, the task of compositional zero-shot trajectory prediction requires developing a model that performs well on both $\mathcal{C^\text{seen}}$ and $\mathcal{C^\text{unseen}}$, despite training only on $\mathcal{C^\text{seen}}$. Unlike conventional CZSL, which requires \textit{predicting labels} for unseen compositions, our setting requires \textit{generating future behavior} for agents in unseen compositional contexts $(c_e, c_s) \in \mathcal{C^\text{unseen}}$.

\begin{figure*}[ht]
  \vspace{0.3cm}
  \centering
  \includegraphics[width=\textwidth, trim=20pt 0pt 0pt 0pt, clip]{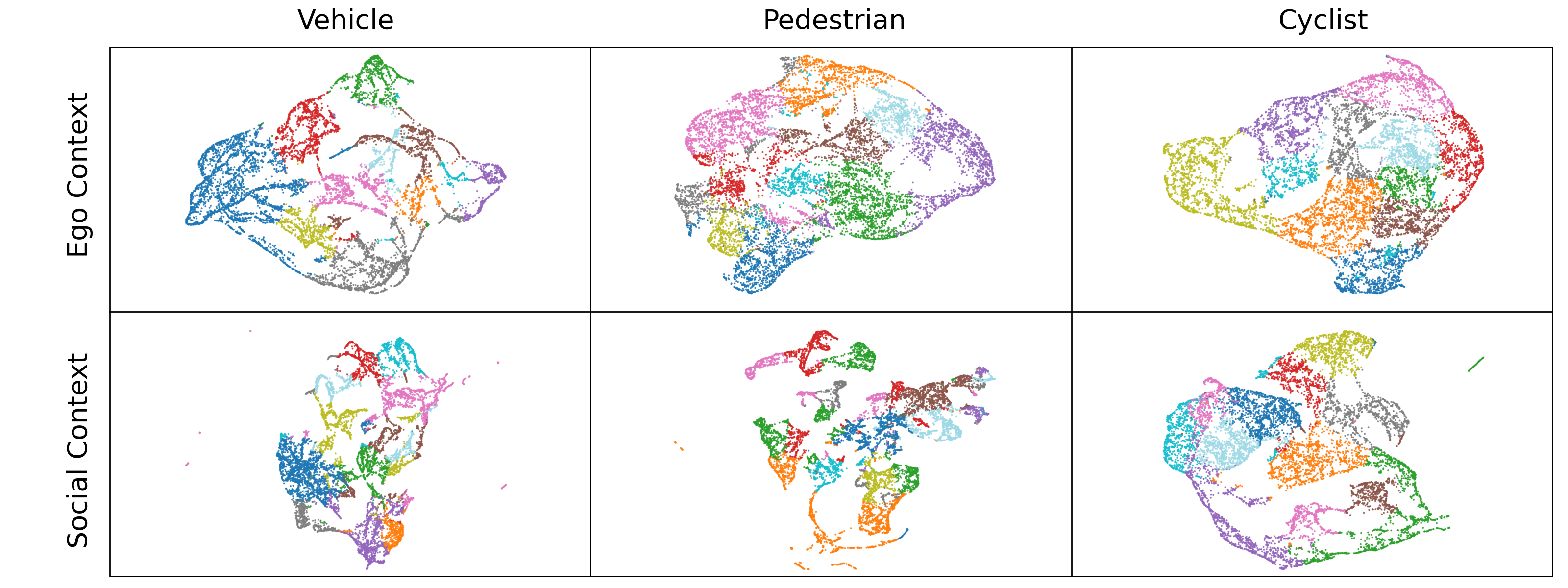}
  \caption{UMAP~\cite{mcinnes2018umap} visualizations for ego and social contexts across agent types. Colors correspond to discretized context labels from clustering described in \Cref{ssec:discretization}, independently per diagram.}
  \label{fig:context_umap}
    \vspace{-0.5cm}

\end{figure*}

\section{Approach} \label{sec:approach}

To both construct long-tail, safety-relevant compositional settings for trajectory prediction and enhance generalization performance therein, we first extract relevant ego and social features (\Cref{ssec:characterization}). We then discretize these features into context labels (\Cref{ssec:discretization}) and use them to create challenging train/test splits (\Cref{ssec:settings}). Finally, we extend baseline trajectory prediction models with new generalization modules to improve out-of-distribution (OOD) performance (\Cref{ssec:generalization}).

\subsection{Safety-Relevant Feature Extraction} \label{ssec:characterization}

We begin by extending prior work on scenario characterization, namely, SafeShift~\cite{stoler2024safeshift} and IMGTP~\cite{ye2024imgtp}, deriving broader sets of both safety-relevant and general attributes and systematically processing them to support downstream discretization. Given a focal agent $i$ in scenario $\mathbf{s}$, we first linearly interpolate $\mathbf{X}_i^\text{hist}$ and each corresponding $\mathbf{X}_{k \neq i}^\text{hist}$ over agent $i$'s valid timestep bounds. We utilize only information from the \textit{history} portion of $\mathbf{s}$ to avoid any leakage from ground-truth future trajectories. We then compute and process the following features:

\noindent \textbf{Ego features:} These focus on kinematic details, lane information, and traffic control device proximity. We first extract agent $i$'s position, velocity, acceleration, and curvature, relative to its final pose (\idest its position and heading at $T_\text{hist}$).  Then, if a valid lane assignment exists based on heading similarity and lateral distance thresholds, we include speed-limit compliance, lane type, and lane-relative, Frenet coordinates of agent $i$'s trajectory. Finally, we obtain distances and relative headings to stop signs, crosswalks, traffic lights, and speed bumps, if such infrastructure is present.

\noindent \textbf{Social features:} These instead focus on agent $i$'s interactions with other relevant agents. We start by computing global scalar attributes, like scenario density. Then, for each non-stationary external agent within a given distance threshold to the focal agent, we compute a \textit{geometry} type of the interaction at $T_\text{hist}$. We first assign interactions into collinear, parallel, opposite, or crossing types, using longitudinal and lateral distances, as well as relative heading differences. We further break down collinear geometries to leading, trailing, or head-on variants, and all other geometries to left or right variants relative to agent $i$.

Next, we compute time-varying kinematic differences to the focal agent, both relative to agent $i$'s final pose as well as to each of $i$'s poses in $\mathbf{X}_i^\text{hist}$. We explicitly capture collision-relevance of the interaction via closing speed and linearly projected conflict points at a fixed future horizon (\idest projected locations of closest separation). For such conflict points, we compute the distance and pose of the point itself, as well as the difference in time-to-conflict-point ($\Delta TTCP$, established in \cite{zhan2019constructing}). Finally, we include a categorical feature capturing the object type (\idest vehicle, pedestrian, or cyclist) of the external agent.

\begin{figure*}[t]
  \vspace{0.3cm}
  \centering
  \includegraphics[width=\textwidth, trim=20pt 0pt 0pt 0pt, clip]{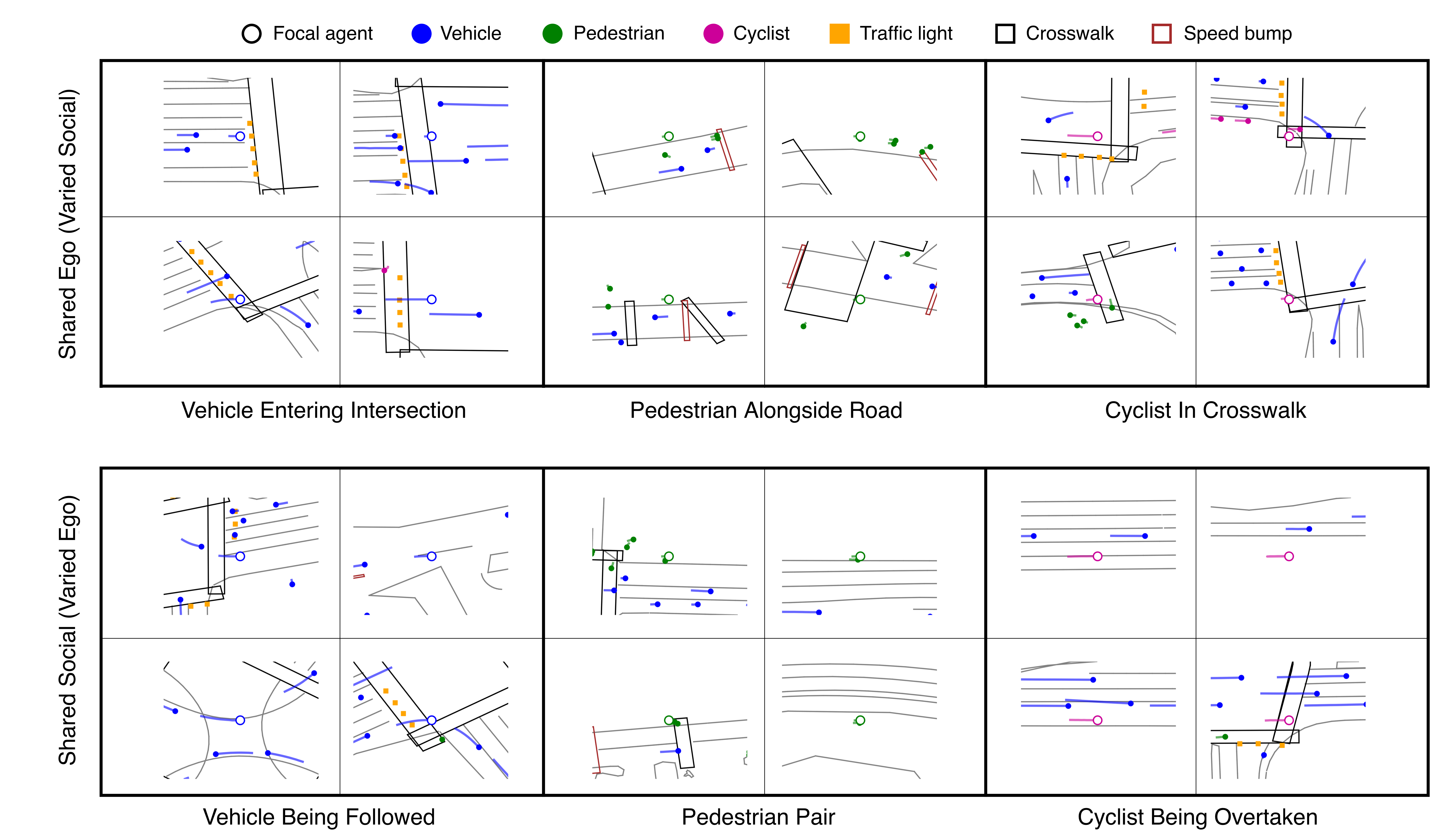}
  \caption{Cluster examples, by context and agent behavior types. In each subgrid of 4 examples, one context (\exempli ego) is fixed while the paired context (\exempli social) varies, with a brief caption describing the shared behavior. Agent markers show positions at $T_\text{hist}$; the focal agent is shown as a large, open circle, while background agents are smaller and filled.}
  \label{fig:centroids}
  \vspace{-0.5cm}
\end{figure*}

\subsection{Feature Processing and Context Discretization}\label{ssec:discretization}

A key challenge in discretizing the above ego and social features into  $(c_e, c_s)$ context labels is that scenarios contain variable numbers and types of agents, traffic control devices, and other scenario elements, even including cases where none are present. We, therefore, process the extracted features into consistent, fixed-length representations, better supporting autoencoding and clustering.

We first organize input features into co-occurring groups. For ego features, these groups include the focal agent's kinematics, lane information, the closest instance of each traffic control device type, and the closest instance of each type located ahead of agent $i$. For social features, these groups include global features and the closest external agent of each geometry type. Within each group, we one-hot encode categorical values, summarize time-varying numerical values with scalar statistics (\idest mean, standard deviation, minimum and maximum, and average slope), and directly include static numerical values.  We then concatenate the features from all groups into a consistently ordered vector for each axis; if any group is absent (\exempli there is no forward stop-sign relative to agent $i$, or no agent trailing agent $i$ collinearly, etc.), we zero-fill the required features. Finally, we append a valid bit to each group indicating whether or not it was present, producing the final ego and social vectors $v_e$ and $v_s$.

Next, we train autoencoders for ego and social vectors separately. We further split by object type and train independent models for each type, allowing distinct latent spaces to be learned for \exempli pedestrian focal agents versus vehicle focal agents. Each autoencoder consists of a simple encoder and decoder multi-layer perceptron (MLP), with layer normalization and dropout on hidden layers; the encoder maps down to a low-dimensional latent space and the decoder maps back to the original feature space. That is, we compute $z = \text{Enc}(v)$ and $\tilde{v} = \text{Dec}(z)$. We train the models primarily with a mean-square error (MSE) reconstruction loss between $v$ and $\tilde{v}$, along with a deep embedding clustering (DEC)~\cite{xie2016unsupervised} loss for regularization on the latent $z$ values.

We then obtain discrete ego and social contexts by performing clustering within the latent spaces captured by these autoencoders, using k-means with $k=11$. We use the Waymo Open Motion Dataset (WOMD)~\cite{ettinger2021large} as a representative source of AD scenarios, sampling approximately 20\% of the total data. To quantitatively assess cluster and latent space coherence, we compute silhouette scores on held-out sets~\cite{shahapure2020cluster}, observing values ranging from $0.31$ to $0.50$, which indicates a reasonably well-structured space. We also visualize UMAP~\cite{mcinnes2018umap} projections of the resulting spaces in \Cref{fig:context_umap}, showing clear separation and evidence of potential sub-clusters. We further present cluster examples for each latent space in \Cref{fig:centroids}, demonstrating intra-cluster consistency alongside paired-context diversity. Ultimately, these intermediate results confirm the validity of our ego and social context derivations, supporting their use as $c_e$ and $c_s$ in downstream compositional tasks.

\subsection{Closed-World and Open-World Settings} \label{ssec:settings}

To ensure that held out context combinations indeed contain sufficient long-tail behaviors, we follow UniTraj~\cite{feng2024unitraj} in considering Kalman difficulty as a reasonably well-calibrated approximation for the overall trajectory prediction challenge. That is, we compute the final displacement error a linear Kalman filter incurs when forecasting the focal agent $i$, averaged at fixed time horizons (\idest 2, 4, and 6 seconds in the future). We then average this difficulty value for each clustered context obtained in \Cref{ssec:discretization}.

Given these context difficulty values, we construct our  \textit{open-world} setting by greedily holding out all agents in the highest average-difficulty ego context (regardless of paired social context), as well as those in the highest average-difficulty social context (regardless of paired ego context), repeating this process until a desired test-set size is achieved. For our \textit{closed-world} setting, we follow the same process, but for a given ego or social context that was held out, we add back in examples from \textit{half} of its co-occurring paired contexts. That is, if \exempli all of $c_e$ was originally held out, we add back $(c_e, c_s^{\{1,2,...,N_\text{social}/2\}})$.

In both settings, the remaining examples are then split uniformly into training and validation sets of desired sizes. Again using WOMD~\cite{ettinger2021large} as a representative dataset, we observe that the average Kalman difficulties in the created test sets are both approximately $20\%$ higher than their corresponding validation set difficulties. This similar magnitude of increase thus allows us to examine the impact of the \textit{compositional} differences between the two long-tail settings, when used in downstream motion prediction experiments.

\subsection{Generalization Strategies} \label{ssec:generalization}

We propose two techniques for improving performance in the above challenging settings: extending task-modular networks (TMNs)~\cite{purushwalkam2019task} for behavior generation, and refining intermediate representations through a difficulty-prediction auxiliary objective. Given a prototypical encoder-decoder trajectory prediction model, we operate directly in its bottleneck layer for both of these ideas. That is, given an intermediate representation of a focal agent as $h$, we transform it into an enriched $h'$ before passing it to the model's decoder.

As in the original TMN, the intuition is that a gating network produces weightings to leverage learned modules in novel ways, as appropriate for novel context combinations. However, whereas the original TMN approach ultimately yields a compatibility score between various candidate context labels and the input representation, we refine the original $h$ representation conditioned on fine-grained, sample-level latents.

Our TMN-inspired architecture thus consists of two main components: a sequence of $N$ layers of $M$ MLP ``modules'', and a latent-conditioned gating MLP $\mathcal{G}$. The first module layer processes $M$ copies of the initial $h$ vector, while modules in subsequent layers process weighted sums of the previous layer's outputs; these weights are obtained from the gating network's output. A linear projection head then maps the final module layer back to the original input space, producing $h'$.

To help ensure that such weightings from $\mathcal{G}$ are coherent, even in open-world settings, we condition $\mathcal{G}$ on the structured latent spaces obtained via the autoencoding process described in \Cref{ssec:discretization}. Note that to avoid information leakage from the test split, we retrain these autoencoders on the closed-world and open-world train sets described in \Cref{ssec:settings}, producing $z_e'$ and $z_s'$ vectors that can safely be used as input to $\mathcal{G}$. We train this component jointly with the baseline prediction model's objectives.

We additionally propose to further refine this $h'$ by enhancing it with difficulty-awareness. We add an auxiliary head, formulated as a simple linear layer on top of $h'$, to map to three scalar values, representing the anticipated Kalman error at 2, 4, and 6 seconds (trained with an MSE regression loss). In this way, we help promote implicit reasoning over difficulty-conditioned decision making (\exempli cautious versus aggressive behavior, etc.), which is especially helpful in our long-tail, safety-informed settings.

\begin{table*}[t]
\vspace{0.3cm}
\caption{Generalization results for the compositional settings; \texttt{Seen} results are in-distribution while \texttt{Unseen} results are out-of-distribution to the train set. Numbers in parentheses indicate relative change from the MTR baseline value in \texttt{Seen} (\idest the setting's generalization gap on that metric). \textbf{Lower} numbers are better for all metric values and relative changes.}
\label{tab:main}
\resizebox{\textwidth}{!}{%

\begin{tabular}{llllllll}
\toprule
Setting & Method & Seen ADE & Seen FDE & Seen Brier-FDE & Unseen ADE & Unseen FDE & Unseen Brier-FDE \\
\midrule
\multirow{4}{*}{Closed-World} & MTR & $2.11\ (-)$ & $4.63\ (-)$ & $5.12\ (-)$ & $2.11\ (+0.2\%)$ & $4.99\ (+7.8\%)$ & $5.47\ (+6.9\%)$ \\
\cmidrule[0.2pt]{2-8}
  & MTR + TMN-Inspired & $2.11\ (-0.1\%)$ & $4.63\ (-0.1\%)$ & $5.10\ (-0.4\%)$ & $2.10\ (-0.6\%)$ & $4.94\ (+6.8\%)$ & $5.42\ (+5.9\%)$ \\
  & MTR + Auxiliary & $2.06\ (-2.6\%)$ & $\mathbf{4.37\ (-5.6\%)}$ & $\mathbf{4.85\ (-5.2\%)}$ & $\mathbf{2.08\ (-1.5\%)}$ & $4.85\ (+4.7\%)$ & $5.33\ (+4.1\%)$ \\
  & MTR + Both & $\mathbf{2.05\ (-2.8\%)}$ & $4.41\ (-4.7\%)$ & $4.89\ (-4.5\%)$ & $2.11\ (+0.1\%)$ & $\mathbf{4.84\ (+4.5\%)}$ & $\mathbf{5.32\ (+3.9\%)}$ \\
\midrule
\multirow{4}{*}{Open-World} & MTR & $1.97\ (-)$ & $4.33\ (-)$ & $4.82\ (-)$ & $2.12\ (+7.8\%)$ & $5.15\ (+19.0\%)$ & $5.65\ (+17.4\%)$ \\
\cmidrule[0.2pt]{2-8}
  & MTR + TMN-Inspired & $2.00\ (+1.7\%)$ & $4.41\ (+1.8\%)$ & $4.89\ (+1.6\%)$ & $2.12\ (+7.5\%)$ & $5.13\ (+18.5\%)$ & $5.62\ (+16.8\%)$ \\
  & MTR + Auxiliary & $2.07\ (+5.1\%)$ & $4.52\ (+4.5\%)$ & $5.01\ (+4.1\%)$ & $2.14\ (+8.6\%)$ & $5.17\ (+19.5\%)$ & $5.67\ (+17.7\%)$ \\
  & MTR + Both & $\mathbf{1.94\ (-1.5\%)}$ & $\mathbf{4.28\ (-1.0\%)}$ & $\mathbf{4.77\ (-1.0\%)}$ & $\mathbf{2.06\ (+4.8\%)}$ & $\mathbf{5.01\ (+15.6\%)}$ & $\mathbf{5.49\ (+14.1\%)}$ \\
\bottomrule
\end{tabular}

}
\vspace{-0.2cm}
\end{table*}

\section{Experiments}
\label{sec:experiments}

\noindent \textbf{Dataset and training details:}
As in our intermediate results in \Cref{sec:approach}, we use the Waymo Open Motion Dataset (WOMD)~\cite{ettinger2021large} as our high quality source of scenarios, processed into a standardized ScenarioNet~\cite{li2023scenarionet} format. We sample approximately 500k agent trajectories from these scenarios to perform ego and social context derivation, evaluation setting creation, and trajectory prediction experiments therein. As in the standard WOMD setting, we set $T_\text{hist}$ to 11 and $T_\text{fut}$ to 80 timesteps, at 10 Hz.

In the characterization process, we use a heading alignment threshold of 30 degrees and a lateral threshold of 6.5m when considering lane assignments. We additionally set an overall distance threshold of 50m for filtering out irrelevant agents, and use relative heading thresholds of 30 degrees and a collinear lateral threshold of 3.25m when determining geometry types. We also use a 10 second horizon when considering linearly projected conflict points.

When training autoencoders for context discretization, we use a latent dimension of $z=16$, compressing and reconstructing $v_e$ and $v_s$ with input dimensions of 346 and 1443, resulting in models with approximately 500k and 800k parameters, respectively. We then use $k=11$ clusters both for the DEC loss module and the k-means computation.

For conducting trajectory prediction experiments, we use the well-established Motion Transformer (MTR)~\cite{shi2022motion} model, scaled to approximately 2.5M total parameters. In our generalization strategies, we utilize $N=3$ intermediate layers and $M=12$ modules per layer, where the gating network conditions on both $z_e'$ and $z_s'$, with dimensions of 16 each. Overall, we add fewer than 100k new parameters to the base MTR model, across both ideas in \Cref{ssec:generalization}. For finer-grained analyses, we train MTR augmented with the TMN-inspired component and auxiliary head jointly (``MTR + Both''), as well as with each component added separately as ablations (``MTR + TMN-Inspired'' and ``MTR + Auxiliary''). We train all models for 30 epochs, with an initial learning rate of $1e-3$, a fixed decay schedule, and a batch size of 192.

\noindent \textbf{Metrics and evaluation:}
We use standard motion forecasting metrics, focusing on displacement differences between ground truth positions in $X^\text{fut}$ and predicted trajectories from the above models. Since future motion is multi-modal, typical models produce multiple distinct future trajectories along with confidence or probability values that a particular mode is best; we set the number of modes to six, as in WOMD~\cite{ettinger2021large} and Argoverse~\cite{wilson2023argoverse}. Then, we utilize the minimum Average and Final Displacement Errors (ADE and FDE) to capture L2 distances over all valid $T_\text{fut}$ timesteps, as well as just at the final observed $T$, for the mode which best matches the ground truth. We additionally report Brier-FDE values, which penalizes FDE by $(1 - p)^2$, where $p$ is the confidence in the best mode, as popularized in the Argoverse challenge. 

Although the closed-world and open-world test sets in \Cref{ssec:settings} contain more examples of difficult, long-tail behavior than their corresponding validation sets, the distribution still remains imbalanced with an over-representation of ``easy'' cases. To better isolate the impacts of the zero-shot compositional challenges, as well as the efficacy of our generalization strategies, we thus report all metrics averaged over the Kalman difficulty-class stratification established in UniTraj~\cite{feng2024unitraj}. Finally, for all experiments, we select the best performing validation Brier-FDE checkpoint and use it to conduct inference on the held-out test sets, obtaining performance measures on both in-distribution compositions in $\mathcal{C^\text{seen}}$ and out-of-distribution compositions in $\mathcal{C^\text{unseen}}$.

\section{Results} \label{sec:results}

We present our main experiment results in \Cref{tab:main}. To understand the severity of our long-tail compositional zero-shot settings, we first report baseline metric values with MTR alone. As in SafeShift~\cite{stoler2024safeshift}, we report each metric along with its relative change from this MTR baseline \texttt{Seen} value; error values increase by an average of $\mathbf{5.0\%}$ and $\mathbf{14.7\%}$ in closed-world and open-world settings, respectively. Note that in both settings, the drop in ADE performance is less than FDE, since shorter $T_\text{fut}$ horizons tend to be easier to predict.

Overall, these results confirm that both our closed-world and open-world zero-shot settings indeed induce significant drops in OOD performance, even for a state-of-the-art model like MTR. Importantly, despite the fact that the relative Kalman difficulty between \texttt{Seen} and \texttt{Unseen} sets is similar in both settings, the open-world drop in performance was \textit{far} larger than the closed-world drop, clearly demonstrating the impact of compositional setting design, as discussed in \Cref{ssec:settings}. 

We then assess our generalization strategies proposed in \Cref{ssec:generalization}. While the TMN-inspired and auxiliary loss components alone improve certain metrics in our settings, using both components achieves the strongest and most consistent performance, highlighting the efficacy of their joint refinement of the hidden representation space. In particular, our full approach reduces the \texttt{Unseen} performance gap to an average of $\mathbf{2.8\%}$ and $\mathbf{11.5\%}$ in closed-world and open-world, respectively. Furthermore, performance in the \texttt{Seen} setting improved by an average of $4.0\%$ in closed-world and $1.2\%$ in open-world. Hence, our proposed strategy of combining modular reasoning using ego and social representations, along with implicitly considering the difficulty of a given example, improves performance broadly, but especially in the OOD \texttt{Unseen} sets.

\section{Conclusion} \label{sec:conclusion}

Robust development and evaluation of trajectory prediction models remain an essential challenge in autonomous driving, especially in the presence of long-tail, safety-critical scenarios encountered in deployment, which are rare or missing altogether in large-scale datasets. We therefore proposed novel evaluation settings on existing datasets, where test sets are constructed to be OOD  with respect to training data on the bases of safety-informed context compositions---to our knowledge, the \textbf{first} extension of the well-established image-labeling CZSL setting to motion prediction. To create these settings, we developed a factorization framework that disentangles traffic participants along ego and social axes, clustering them into paired context labels. We then used these contexts to build long-tail, zero-shot closed-world and open-world settings. On a state-of-the-art baseline model, we observed performance drops from ID data of 5.0\% and 14.7\%, respectively; by extending task-modular gating networks and incorporating an auxiliary, difficulty-prediction loss, we reduced these OOD gaps to 2.8\% and 11.5\%, while also improving ID performance by 4.0\% and 1.2\%, respectively.

Despite the efficacy of our evaluation settings and generalization strategies, further improvements are still possible. In particular, extending other non CLIP-based approaches from image-labeling CZSL, such as learned attention propagation~\cite{khan2023learning}, could boost both ID and OOD performance. Additionally, factorizing along more than two safety-relevant, semantic axes could also increase interpretability, as well as the effectiveness of generalization approaches. We encourage future work to explore these directions.


\footnotesize
\bibliographystyle{IEEEtranBST/IEEEtran}
\balance
\bibliography{IEEEtranBST/IEEEabrv, ref}
\end{document}